\documentclass[letterpaper, 10pt, conference]{ieeeconf}

\IEEEoverridecommandlockouts
\usepackage{amsmath}
\usepackage{amssymb}
\usepackage{graphicx}
\usepackage{booktabs}
\usepackage{hyperref}
\usepackage{color}
\usepackage{multirow}
\usepackage{subcaption}
\usepackage{wrapfig,algorithm,amssymb,algpseudocode}
\usepackage{multirow}
\usepackage{multicol}
\usepackage[table]{xcolor}
\definecolor{myblue}{rgb}{0.88, 0.95, 0.99}
\definecolor{bluecomm}{rgb}{0.25, 0.4, 0.55}
\usepackage{wrapfig,algorithm,amssymb,algpseudocode}
\let\labelindent\relax
\usepackage{enumitem}

\title{\LARGE \bf
GesVLA: Gesture-Aware Vision-Language-Action Model \\with Embedded Representations
}

\author{Wenxuan Guo\textsuperscript{1$*$} \quad
Ziyuan Li\textsuperscript{1$*$} \quad
Meng Zhang\textsuperscript{2$\dagger$} \quad
Yichen Liu\textsuperscript{1} \quad
Yimeng Dong\textsuperscript{1} \quad
Chuxi Xu\textsuperscript{1} \\
Yunfei Wei\textsuperscript{2} \quad
Ze Chen\textsuperscript{2} \quad
Erjin Zhou\textsuperscript{2} \quad
Jianjiang Feng\textsuperscript{1$\ddagger$} \\
	\textsuperscript{1}Tsinghua University \qquad \textsuperscript{2}Dexmal}

\begin{document}
\newcommand{\name}{GesVLA}
\maketitle
{
        \renewcommand{\thefootnote}{* \,}
	\footnotetext[5]{Equal contribution. $^{\dagger}$ Project leader. $^{\ddagger}$ Corresponding author.}
}

\thispagestyle{empty}
\pagestyle{empty}

\begin{abstract}
Vision-Language-Action (VLA) models have shown strong potential for general-purpose robot manipulation by unifying perception and action. However, existing VLA systems primarily rely on textual instructions and struggle to resolve spatial ambiguity in complex scenes with multiple similar objects. To address this limitation, we introduce gesture as a parallel instruction modality and propose a Gesture-aware Vision-Language-Action model (GesVLA). Our approach encodes gesture features directly into the latent space, enabling them to participate in both high-level reasoning and low-level action generation, and adopts a dual-VLM architecture to achieve tight coupling between gesture representations and action policies. At the data level, we construct a scalable gesture data generation pipeline by rendering hand models onto real-world scene images. This reduces the sim-to-real visual gap while producing rich data with diverse motion patterns and corresponding pointing annotations.
In addition, we employ a two-stage training strategy to equip the model with both gesture perception and action prediction capabilities. We evaluate our approach on multiple real-world robotic tasks, including a controlled block manipulation task for validation and more practical scenarios such as product and produce selection. Experimental results show that incorporating gesture consistently improves target grounding accuracy and human-robot interaction efficiency, especially in complex and cluttered environments. 
Project page: \href{https://gwxuan.github.io/GesVLA/}{https://gwxuan.github.io/GesVLA/}.
\end{abstract}

\section{INTRODUCTION}

\begin{figure}[t]
    \centering
    \includegraphics[width=0.98\linewidth]{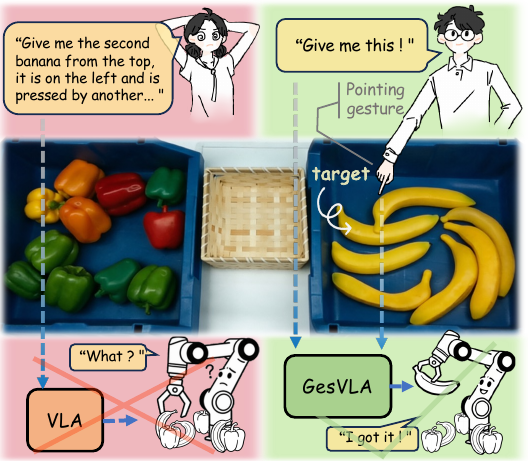}
    \caption{
    Comparison between language-only VLA and our gesture-aware GesVLA. 
    \textbf{Left}: Text-only instructions can be ambiguous in cluttered scenes, requiring complex descriptions. 
    \textbf{Right}: By incorporating pointing gestures, GesVLA enables direct and intuitive target specification, leading to more accurate and efficient task execution.
    }
    \label{fig:teaser}
    \vspace{-.3cm}
\end{figure}

Vision-Language-Action (VLA) models have recently emerged as a promising paradigm for general-purpose robot manipulation~\cite{zitkovich2023rt,kim24openvla,intelligence2025pi_}, enabling robots to interpret human instructions and generate actions in a unified framework. By leveraging large-scale vision-language pretraining~\cite{beyer2024paligemma,black2024pi_0,yu2026dm0}, these models exhibit strong capabilities in instruction following and open-world interaction. However, current VLA systems primarily rely on textual instructions as the main interface for human intent, which introduces inherent limitations in real-world scenarios.

In many practical settings, language alone is insufficient to precisely specify target objects, especially in cluttered environments with multiple similar instances, as illustrated in Fig.~\ref{fig:teaser}. Instructions such as “pick up this one” or “place it there” are inherently ambiguous without additional spatial cues. Humans naturally resolve such ambiguity through deictic gestures, such as pointing, which provide direct and intuitive spatial grounding. Despite advances in hand pose estimation~\cite{zimmermann2019freihand,zhang2020mediapipe}, existing works~\cite{edge2023diver,kobzarev2025gestllm,sassali2025evaluating} typically treat gestures as auxiliary signals or post-processing inputs, rather than integrating them into the core reasoning process of VLA models.
Another critical limitation lies in the lack of suitable training data for gesture-grounded interaction. Existing robotic datasets rarely include aligned gesture, language, and action signals with precise spatial annotations. Collecting real-world gesture data at scale is expensive and difficult to annotate accurately, especially for pointing targets. This data bottleneck limits the ability of VLA models to learn fine-grained gesture understanding.

To address these challenges, we propose a Gesture-aware VLA (GesVLA), which introduces gesture as a first-class modality alongside vision and language, and is supported by a scalable gesture dataset construction pipeline. Instead of converting gestures into discrete textual representations, our approach encodes gesture features directly into continuous latent tokens, enabling seamless multimodal fusion. 

Specifically, we design a dual-VLM architecture that decouples intent reasoning and action generation while maintaining a direct latent interaction between gesture representations and the action policy. Gesture features extracted from hand keypoints are projected into the shared embedding space and jointly modeled with visual and linguistic inputs. Furthermore, we adopt flow-based action generation to produce smooth and precise trajectories conditioned on multimodal context.
To enable effective training, we construct a scalable gesture data generation pipeline by rendering hand models onto real-world scene images with precise pointing target annotations. By grounding synthetic gestures in real visual backgrounds, the generated data exhibits a reduced sim-to-real visual gap. At the same time, the pipeline allows for diverse variations in hand appearance and motion patterns, ensuring strong scalability and coverage. This design provides high-quality and abundant supervision for learning fine-grained gesture understanding.
In addition, we propose a two-stage training strategy that facilitates learning of both gesture perception and action prediction.

We evaluate on multiple real-world tasks, including a controlled block manipulation task, as well as more practical scenarios such as product and produce selection. The results show that incorporating gesture significantly improves target grounding accuracy and task success rates, with more pronounced gains in complex and cluttered environments.

Our contributions can be summarized as follows:
\begin{itemize}
    \item We introduce gesture as a parallel instruction modality in VLA models, and systematically study its role in resolving spatial ambiguity in human-robot interaction.
    \item We design a dual-VLM architecture with gesture-aware latent encoding, enabling tight coupling between gesture representations and action generation.
    \item We construct a scalable gesture data generation pipeline with precise spatial annotations and adopt a two-stage training strategy to support effective model training.
\end{itemize}
\section{RELATED WORK}

\textbf{Vision-Language-Action Models.} Recent years have seen significant progress in robotic manipulation with Vision-Language-Action (VLA) models~\cite{zitkovich2023rt,kim24openvla,black2024pi_0,intelligence2025pi_,lin2025onetwovla,yu2026dm0}. These models perform end-to-end mapping from visual observations to action outputs by unifying perception, language, and action. Some adopt a hierarchical architecture with high-level planning for task decomposition and low-level policy for execution~\cite{han2024dual,bjorck2025gr00t,shi2025hi}, further enhancing VLA’s capability for complex tasks. However, these models rely on a single input modality—language instructions—which often fails to accurately convey task goals. While some works introduce additional input modalities beyond language, such as depth~\cite{yuan2025depthvla}, point clouds~\cite{li2026pointvla}, or touch~\cite{zhang2025vtla}, these only enhance environmental perception during execution, not task understanding. In contrast, we introduce gesture as a new modality that works with language to provide fine-grained spatial reference, fundamentally addressing the challenge of task target localization in complex scenes.

\textbf{Gesture Interaction in Robotics.}
Researchers have integrated gesture perception into robot control to overcome the limitations of pure language. One category relies on traditional geometric rules and ray casting~\cite{edge2023diver, sassali2025evaluating, hu2022augmented}, determining targets via predefined pointing vectors (e.g., shoulder-wrist alignment). While intuitive, they lack robustness and deep language coupling. Another category uses separate modules to convert gestures into text for LLM-based planning~\cite{kobzarev2025gestllm, lin2023gesture}, causing information loss in precision tasks. Others train specialized models to predict pointing regions~\cite{bamani2023recognition, matuszek2014learning, muller2025pointing} or use general VLMs to generate visual prompts for VLA~\cite{yu2025point}. However, these decoupled, multi-module pipelines often suffer from cumulative errors. Our unified architecture treats gesture as a primary modality, employing deep feature-level fusion to maximize the utilization of rich gestural spatial information.
\section{PROBLEM FORMULATION}
We study a gesture-augmented vision-language-action (VLA) problem, where a robot is required to execute manipulation tasks based on multimodal human instructions.
In standard VLA settings, the robot receives visual observations $\mathcal{O}$ and a language instruction $\mathcal{T}$, and aims to predict a sequence of actions that accomplish the task. However, in many real-world scenarios, language alone is insufficient to uniquely specify the target due to spatial ambiguity (e.g., multiple similar objects). To address this limitation, we introduce gesture as an additional instruction modality.

Formally, the input consists of three components: visual observations $\mathcal{O}$, a language instruction $\mathcal{T}$, and a gesture video $\mathcal{G}$. The visual observations $\mathcal{O}$ include multi-view RGB images of the scene, the language instruction $\mathcal{T}$ describes the task at a high level, and the gesture signal $\mathcal{G}$ provides spatial grounding through pointing.
The goal is to predict a sequence of robot actions $\{a_t\}_{t=1}^K$ that successfully accomplish the task. This requires the model to (1) infer the intended target objects or locations from ambiguous multimodal inputs, and (2) generate feasible action trajectories conditioned on both perception and inferred intention.

\begin{figure*}[t]
    \centering
    \includegraphics[width=\linewidth]{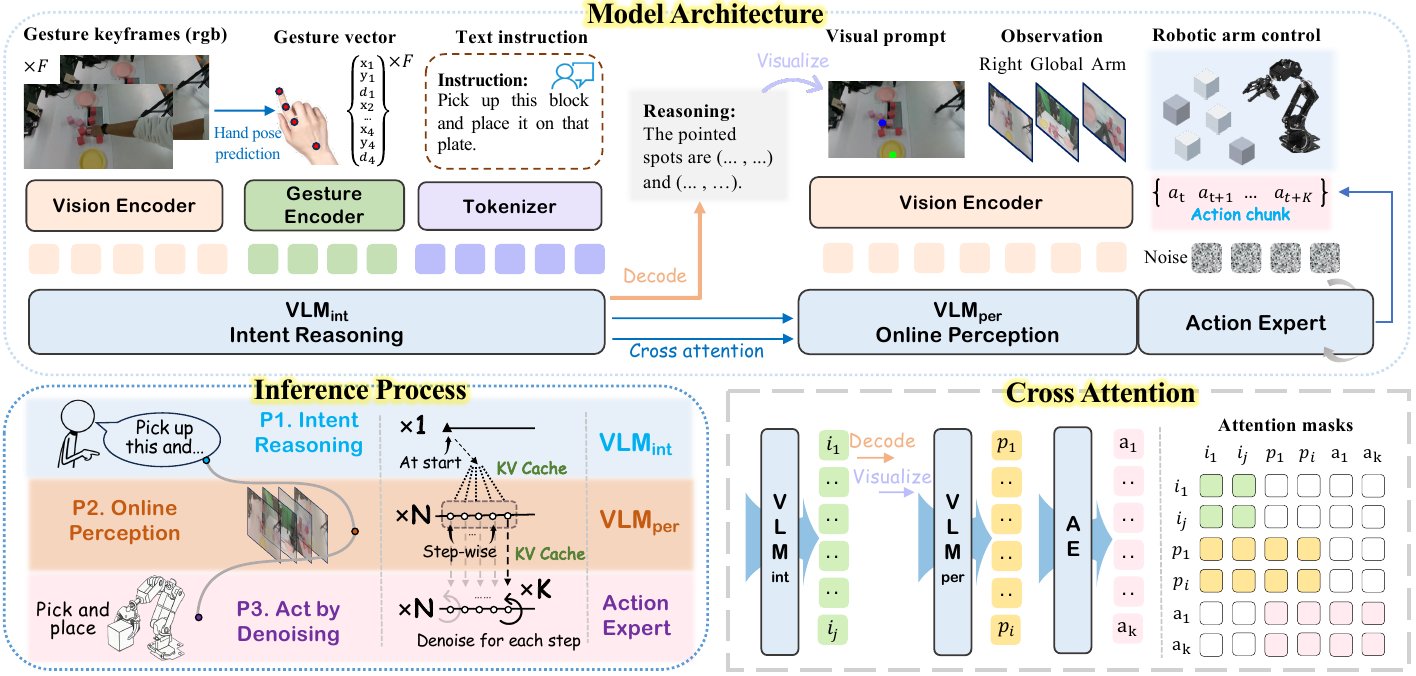}
    \caption{
    Overview of the proposed gesture-aware dual-VLM architecture. 
    $\text{VLM}_{\text{int}}$ performs gesture-conditioned intent reasoning from multimodal instructions, producing textual and visual reasoning outputs. 
    $\text{VLM}_{\text{per}}$ performs online perception by attending to the latent representations of $\text{VLM}_{\text{int}}$ via cross-attention. 
    The action expert generates action trajectories through iterative denoising. 
    The asymmetric attention design enables efficient inference by enforcing a unidirectional information flow, where $\text{VLM}_{\text{int}}$ does not attend to downstream modules, allowing it to be computed once and reused throughout execution.
    }
    \label{fig:architecture}
    \vspace{-.1cm}
\end{figure*}

\section{METHOD}
\subsection{Gesture-aware Dual-VLM Architecture}
As shown in Fig.~\ref{fig:architecture}, we propose a gesture-aware dual-VLM architecture that separates intent reasoning from action generation while enabling tight interaction between them.
The model consists of a VLM for intent reasoning ($\text{VLM}_{\text{int}}$), a VLM for online perception ($\text{VLM}_{\text{per}}$), and an action expert ($\mathcal{F}_{\theta}$).
$\text{VLM}_{\text{int}}$ is responsible for interpreting user intention from gesture and language, while $\text{VLM}_{\text{per}}$ and $\mathcal{F}_{\theta}$ generate actions conditioned on observations and inferred intention.

\textbf{Gesture Embedding.}
The gesture instruction is represented as a sequence of keyframes $\{g_i\}_{i=1}^F$. Instead of processing the full gesture video, we extract keyframes based on hand motion dynamics. Specifically, we empirically observe that users typically exhibit a brief pause when pointing to a target object or location. Based on this, we track hand keypoints over time and select frames corresponding to such motion stagnation as keyframes. This design filters out redundant temporal information and significantly reduces computational cost while preserving essential spatial cues.
For each keyframe, we extract four keypoints using MediaPipe (the wrist and three joints of the index finger), each represented as $(x,y,d)$. These are concatenated into a feature vector $\mathbf{h}_i \in \mathbb{R}^{12}$ and projected into the latent space:
\begin{equation}
\mathbf{z}_i^{g} = \phi(\text{Pose}(g_i)), \quad g_i \in \mathcal{S}(\text{Pose}(\mathcal{G})),
\end{equation}
where $\text{Pose}(\cdot)$ denotes hand pose estimation, $\phi(\cdot)$ is the gesture encoding module composed of multiple MLP layers, and $\mathcal{S}(\cdot)$ denotes the keyframe selection function based on hand motion dynamics.

\textbf{Dual-VLM Architecture.}
We adopt distinct VLMs that separates intention reasoning from action generation while enabling tight interaction between the two stages. 
The intent reasoning model $\text{VLM}_{\text{int}}$ takes gesture inputs $\mathcal{G}$ together with the language instruction $\mathcal{T}$, and infers task-relevant targets:
\begin{equation}
(\mathbf{y}, \mathcal{K}^{\text{int}}, \mathcal{V}^{\text{int}}) = \text{VLM}_{\text{int}}(\mathcal{G}, \mathcal{T}),
\end{equation}
where $\mathbf{y}$ denotes the reasoning outputs, including textual descriptions of the inferred targets, and visual prompts obtained by post-processing and visualizing these outputs to indicate target objects and locations, and $(\mathcal{K}^{\text{int}}, \mathcal{V}^{\text{int}})$ are the cached key-value states across layers.

The online perception model $\text{VLM}_{\text{per}}$ then takes scene observations $\mathcal{O}$ and language instruction $\mathcal{T}$, together with the reasoning outputs $\mathbf{y}$ from $\text{VLM}_{\text{int}}$, to produce a latent representation for action prediction:
\begin{equation}
(\mathcal{K}^{\text{per}}, \mathcal{V}^{\text{per}}) =
\text{VLM}_{\text{per}}(\mathcal{O}, \mathcal{T}, \mathbf{y}; \mathcal{K}^{\text{int}}, \mathcal{V}^{\text{int}}),
\end{equation}
where cross-attention enables $\text{VLM}_{\text{per}}$ to directly attend to the latent representations of $\text{VLM}_{\text{int}}$, allowing intention information to be propagated without intermediate discretization.

Finally, the action expert generates an action chunk conditioned on the latent representations from $\text{VLM}_{\text{per}}$, the current robot state $\mathbf{s}_t$, and an initial action noise $\mathbf{x}_0$:
\begin{equation}
\mathbf{a}_{1:K} = \mathcal{F}_{\theta}(\mathbf{x}_0,\mathbf{s}_t,\mathcal{K}^{\text{per}}, \mathcal{V}^{\text{per}}),
\end{equation}
where $\mathbf{x}_0 \sim \mathcal{N}(\mathbf{0}, \mathbf{I})$ is the initial noisy action trajectory, $\mathbf{s}_t$ denotes the robot state, and $\mathcal{F}_{\theta}$ is a flow-based policy that iteratively denoises $\mathbf{x}_0$ into a continuous action sequence.

\textbf{Asymmetric Interaction and Efficiency.}
As illustrated by the attention mask in Fig.~\ref{fig:architecture}, the interaction between modules is designed to be asymmetric:
\begin{equation}
\text{VLM}_{\text{int}} \rightarrow \text{VLM}_{\text{per}} \rightarrow \text{Action Expert},
\end{equation}
where $\text{VLM}_{\text{int}}$ does not attend to downstream modules. Such a design enables efficient reuse of computation during inference. Specifically, the key-value states $(\mathcal{K}^{\text{int}}, \mathcal{V}^{\text{int}})$ are computed once and reused, while $\text{VLM}_{\text{per}}$ is executed at each control step, and the action expert performs multiple denoising iterations.
This results in a computation ratio of $1:T:T \times N$, where $T$ denotes action prediction steps during execution and $N$ denotes denoising steps. The efficient inference procedure is summarized in Algorithm~\ref{algo:infer}.

\begin{algorithm}[t]
\caption{Inference Procedure of GesVLA}
\label{algo:infer}
\begin{algorithmic}[1]

\State \textbf{Input:} $\mathcal{O}_0, \mathcal{T}, \mathcal{G}, \mathbf{s}_0$
\State Estimate hand poses $\text{Pose}(\mathcal{G})$ and select keyframes $\{g_i\}_{i=1}^F \leftarrow \mathcal{S}(\text{Pose}(\mathcal{G}))$
\State Encode gesture embeddings $\mathbf{z}_i^{g} \leftarrow \phi(\text{Pose}(g_i))$
\State $(\mathbf{y}, \mathcal{K}^{\text{int}}, \mathcal{V}^{\text{int}}) \leftarrow \text{VLM}_{\text{int}}(\mathcal{G}, \mathcal{T})$
\State $\mathcal{O} \leftarrow \mathcal{O}_0,\quad \mathbf{s}_t \leftarrow \mathbf{s}_0$
\While{task not finished}
    \State $(\mathcal{K}^{\text{per}}, \mathcal{V}^{\text{per}}) \leftarrow 
    \text{VLM}_{\text{per}}(\mathcal{O}, \mathcal{T}, \mathbf{y}; \mathcal{K}^{\text{int}}, \mathcal{V}^{\text{int}})$
    \State Initialize action noise $\mathbf{x}_0 \sim \mathcal{N}(\mathbf{0}, \mathbf{I})$
    \State Predict action chunk $\mathbf{a}_{1:K} \leftarrow \mathcal{F}_{\theta}(\mathbf{x}_0, \mathbf{s}_t, \mathcal{K}^{\text{per}}, \mathcal{V}^{\text{per}})$
    \State Execute the action chunk $\mathbf{a}_{1:K}$ (or part: $\mathbf{a}_{1:k}$)
    \State Update observation $\mathcal{O}$ and robot state $\mathbf{s}_t$
\EndWhile
\end{algorithmic}
\end{algorithm}

\begin{figure*}[t]
    \centering
    \includegraphics[width=\linewidth]{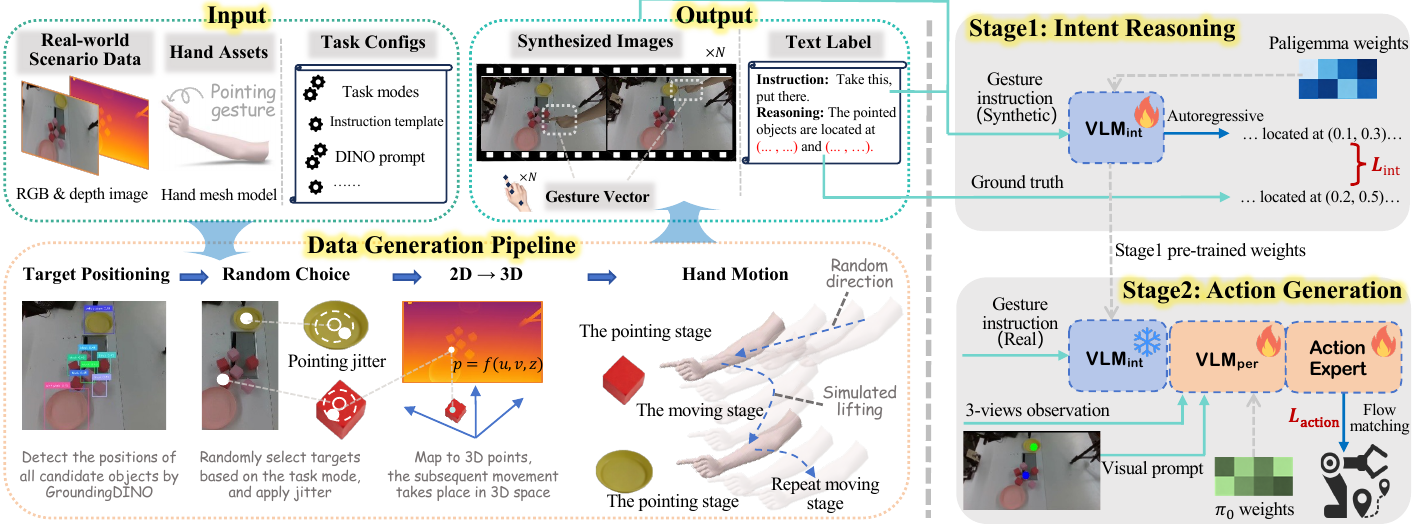}
    \caption{
    Scalable gesture-aware data engine and two-stage training pipeline of GesVLA. 
    \textbf{Left}: The semi-synthetic data engine synthesizes gesture trajectories on real RGB-D scenes, producing gesture data and paired language instructions with precise spatial annotations. 
    \textbf{Right}: In Stage 1, $\text{VLM}_{\text{int}}$ is trained for gesture-conditioned intent reasoning using the generated data. In Stage 2, the dual-VLM policy is trained on real robot demonstrations with flow matching for action generation.
    }
    \label{fig:data_training}
    \vspace{-.1cm}
\end{figure*}

\subsection{Scalable Gesture-aware Data Engine}

A key challenge in training gesture-aware embodied models is the lack of large-scale data with accurate pointing supervision. Collecting such data in real scenes is expensive, while manual annotation of pointing targets is inherently ambiguous. To address this issue, we construct a scalable gesture data generation pipeline that synthesizes gesture trajectories together with corresponding language instructions. The generation pipeline is shown in Fig.~\ref{fig:data_training}.

\textbf{Target Selection and 3D Grounding.}
Given a real scene image and its depth map, we first detect candidate objects using GroundingDINO~\cite{liu2023grounding}. We then randomly sample task-relevant targets from the detected candidates. For the selected target box, we take its center with random jitter to simulate natural pointing deviation, and recover the corresponding 3D target point using camera intrinsics:
\begin{equation}
\mathbf{p}=\left(\frac{(u-c_x)z}{f_x},\ \frac{(v-c_y)z}{f_y},\ z\right),
\end{equation}
where $(u,v)$ is the jittered image coordinate, $z$ is the depth value at that location, and $(f_x,f_y,c_x,c_y)$ are the camera intrinsic parameters. This step provides an accurate 3D supervision anchor for subsequent hand motion generation. The same procedure is applied to multiple targets, enabling diverse task patterns such as picking up arbitrary numbers of objects and placing them at different locations.

\textbf{Pointing Motion Generation.}
To simulate the natural process of pointing at an object, the hand is initialized from a random direction and gradually moves toward the target point. Let $\mathbf{p}_t$ denote the target point and $\mathbf{d}$ be a sampled unit direction. At step $k$, the hand position is defined as:
\begin{equation}
\mathbf{p}_h = \mathbf{p}_t - \mathbf{d}\,(\tau + k\Delta),
\end{equation}
where $\tau$ is a stopping threshold and $\Delta$ is the step size. This produces a smooth approach trajectory that ends near the target while preserving a valid pointing pose. A hand mesh is then transformed and rendered at each step to produce the gesture video frames.

Our tasks typically involve multiple pointing targets, requiring the hand to move from one referenced location to another. Starting from the current hand position, we rotate the pointing direction toward the next target and update the visible hand trajectory with a parabolic lifting term:
\begin{equation}
\mathbf{p}_{\mathrm{vis}}=\mathbf{p}_h+h_{\max}\bigl(1-(2\alpha-1)^2\bigr)\mathbf{n}_{\mathrm{up}},
\end{equation}
where $\alpha\in[0,1]$ is the normalized motion progress, $h_{\max}$ controls the maximum lift height, and $\mathbf{n}_{\mathrm{up}}$ is the upward direction. This term makes the generated motion more human-like than straight-line interpolation.

\textbf{Multimodal Dataset.}
Following the above pipeline, we construct a semi-synthetic gesture dataset containing approximately 16k samples. Each sample consists of multiple modalities and annotations for training and evaluation.
Specifically, each sample includes: 
(1) a video $\mathcal{O}$ capturing the pointing process, where the camera viewpoint is consistent with the original real-scene observation; 
(2) a language instruction $\mathcal{T}$ describing the task (e.g., ``pick up this and put it there''); 
(3) supervision signals $\mathbf{y}$ for intent reasoning model $\text{VLM}_{\text{int}}$, including targets and locations; 
and (4) metadata, such as task type indices and original scene indices.

\textbf{Discussion.}
The above pipeline directly combines real-scene backgrounds with controllable synthetic hand motion. It preserves realistic scene appearance while providing exact pointing labels at scale. Since only scene images, depth maps, hand assets, and task configuration are required, the pipeline can be easily extended to new objects, new tasks, and new motion patterns without additional manual annotation. This makes it particularly suitable for pre-training gesture reasoning models with rich spatial supervision.

\subsection{Training Pipeline of GesVLA}
GesVLA is trained in two stages, as shown in Fig.~\ref{fig:data_training}. In the first stage, $\text{VLM}_{\text{int}}$ is trained on the large-scale semi-synthetic gesture dataset to learn gesture-conditioned spatial reasoning. In the second stage, the full policy is trained on real robot demonstrations, where $\text{VLM}_{\text{int}}$ is frozen and only $\text{VLM}_{\text{per}}$ and the action expert $\mathcal{F}_{\theta}$ are optimized.

\textbf{Intent Reasoning Pre-training.}
In this stage, $\text{VLM}_{\text{int}}$ is trained using the semi-synthetic multimodal dataset. Each sample contains gesture video (represented by keyframes), hand keypoint features, and a language instruction $\mathcal{T}$, together with supervision on reasoning outputs $\mathbf{y}$.

To make spatial prediction compatible with autoregressive generation, target coordinates are discretized into a fixed number of bins and represented as special tokens. The model is trained to generate both textual reasoning and coordinate tokens in a unified sequence.
The training objective is standard teacher-forced autoregressive cross-entropy:
\begin{equation}
\mathcal{L}_{\text{int}}
=
-\frac{1}{M}\sum_{i=1}^{M}
\log P(a_i^{*}\mid a_{<i}^{*}, \mathcal{G}, \mathcal{T}),
\end{equation}
where $\{a_i\}$ includes both text tokens and discretized coordinate tokens. It allows $\text{VLM}_{\text{int}}$ to jointly learn semantic reasoning and spatial grounding in a unified token space.

\textbf{Joint Training for Action Generation.}
In the second stage, we train the dual-VLM policy on real robot demonstrations by sampling training tuples at the step level. Specifically, we first sample a multimodal instruction, which includes gesture keyframes, preprocessed hand keypoints, and a language instruction $\mathcal{T}$. 
Given this, we then sample a time step from the corresponding manipulation trajectory, and take the multi-view observations at that step as input to $\text{VLM}_{\text{per}}$. 
Meanwhile, a random denoising step $t$ is sampled to construct the noisy action input for the action expert.

The action expert is trained to predict the denoising direction conditioned on the latent representation from $\text{VLM}_{\text{per}}$ and the robot state, enabling iterative refinement of action trajectories.
We adopt a flow matching objective to model continuous action distributions:
\begin{equation}
\mathcal{L}_{\text{action}}
=
\mathbb{E}_{\mathbf{x}_{0},\mathbf{x}_{1},t}
\left[
\left\|
\mathbf{v}_{\theta}(\mathbf{x}_{t}, t, \mathbf{c})
-
(\mathbf{x}_{1}-\mathbf{x}_{0})
\right\|^{2}
\right],
\end{equation}
where $\mathbf{c}$ denotes the conditioning context from $\text{VLM}_{\text{per}}$ and the robot state, and $\mathbf{v}_{\theta}$ represents the velocity prediction at a single step in the multi-step denoising process.

\subsection{Implementation Details}
We employ the PaliGemma-2B architecture~\cite{beyer2024paligemma} as the backbone of $\text{VLM}_{\text{int}}$ and $\text{VLM}_{\text{per}}$. 
In the first stage, we perform intent reasoning pre-training, where $\text{VLM}_{\text{int}}$ is initialized from PaliGemma~\cite{beyer2024paligemma}. In the second stage, we train the policy on real robot data, with $\text{VLM}_{\text{int}}$ frozen and $\text{VLM}_{\text{per}}$ initialized from the VLM backbone of $\pi_0$~\cite{black2024pi_0}, while the action expert is trained from scratch. 
Both stages are optimized using AdamW with a warmup cosine learning rate schedule and exponential moving average (EMA) of parameters.
Training is conducted on $4$ H20 GPUs, while inference is performed on a single RTX 4090 GPU.

\section{EXPERIMENTS}
\begin{figure*}[t]
    \centering
    \includegraphics[width=\linewidth]{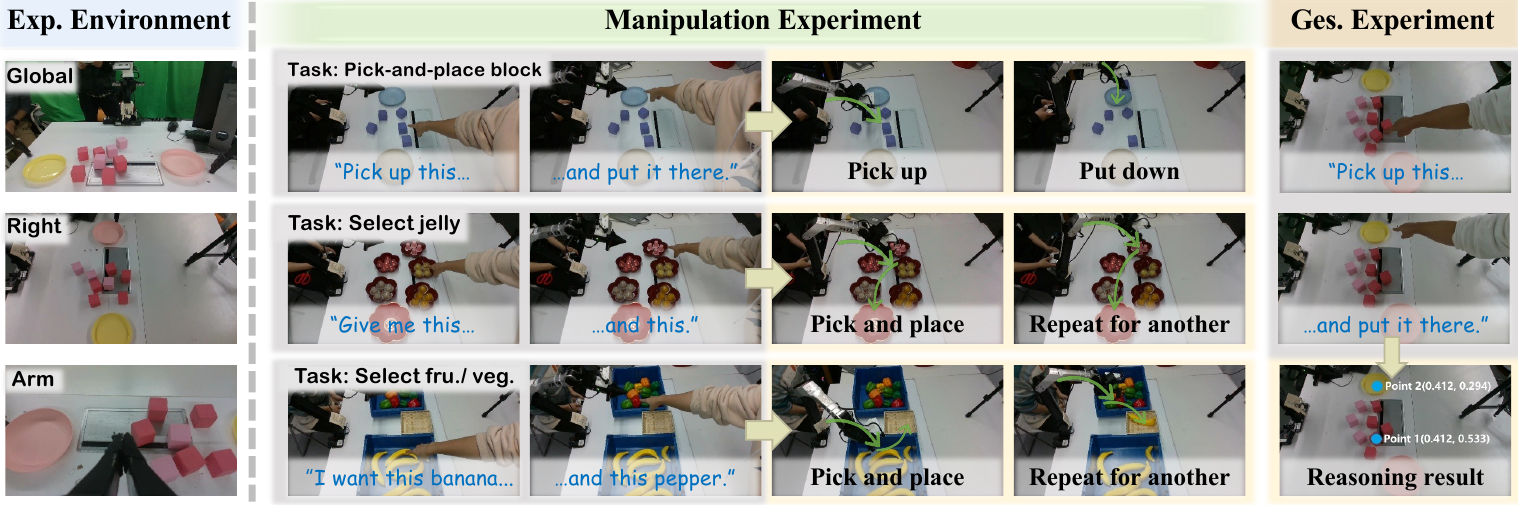}
    \caption{
    Experimental setup and tasks in real-world environments. 
    The manipulation tasks include Pick-and-Place Block, Select Jelly, and Select Fruit/Vegetable under multimodal gesture-language instructions.
    We also independently evaluate gesture-conditioned intent reasoning tasks, where the model predicts target objects from pointing gestures.
    }
    \label{fig:setup}
\end{figure*}
\subsection{Experimental Setup}\label{exp_setup}
The experimental setup is shown in Fig.~\ref{fig:setup}. We conduct both intent reasoning and robot manipulation experiments. Three cameras with a frame rate of 30 fps are deployed on a fixed table: two provide a global view and a right-side view, respectively, and the third is mounted on the robotic gripper.

\begin{table*}[t]
    \small
    \centering
    \caption{Real-robot manipulation experiment results. For Pick-and-Place Block, simple/hard subsets are divided by the number of blocks on the table: simple contains fewer than 5 blocks, while hard contains 5 or more blocks. For Select Jelly and Select Fruit/Vegetable, simple denotes single-object pick-and-place tasks, while hard denotes sequential multi-object pick-and-place tasks. Avg values are success rates over 20 trials per task. Total Avg is the mean of the three task averages.}
    \vspace{-4pt}
    \setlength{\tabcolsep}{6pt}
    \scalebox{0.99}{
    {\fontsize{8.9pt}{13pt}\selectfont
    \begin{tabular}{lcccccccccc}
    \toprule
    \multirow{2}{*}{Method} & \multicolumn{3}{c}{Pick-and-Place Block} & \multicolumn{3}{c}{Select Jelly} & \multicolumn{3}{c}{Select Fruit and Vegetable} & \multirow{2}{*}{Avg (\%)} \\
    \cmidrule(lr){2-4} \cmidrule(lr){5-7} \cmidrule(lr){8-10}
    & Simple & Hard & Avg (\%) & Simple & Hard & Avg (\%) & Simple & Hard & Avg (\%) & \\
    \midrule
    Text-only VLA~\cite{black2024pi_0} & 6/10 & 3/10 & 45.0 & 4/10 & 3/10 & 35.0 & 2/10 & 1/10 & 15.0 & 31.7 \\
    MLLM~\cite{yang2025qwen3} + VLA~\cite{black2024pi_0} & 6/10 & 4/10 & 50.0 & 4/10 & 2/10 & 30.0 & 2/10 & 1/10 & 15.0 & 31.7 \\
    Geometric pipeline + VLA~\cite{black2024pi_0} & 7/10 & 4/10 & 55.0 & 3/10 & 3/10 & 30.0 & 4/10 & 4/10 & 40.0 & 41.7 \\
    \rowcolor{myblue}
    GesVLA (decoupled VLMs) & \textbf{10/10} & 4/10 & 70.0 & 8/10 & 3/10 & 55.0 & 7/10 & 5/10 & 60.0 & 61.7 \\
    \rowcolor{myblue}
    \textbf{GesVLA (ours)} & \textbf{10/10} & \textbf{9/10} & \textbf{95.0} & \textbf{9/10} & \textbf{6/10} & \textbf{75.0} & \textbf{8/10} & \textbf{8/10} & \textbf{80.0} & \textbf{83.3} \\
    \bottomrule
    \end{tabular}}}
    \vspace{-.2cm}
    \label{tab:real_robot}
\end{table*}

\begin{table}[t]
    \small
    \centering
    \caption{Quantitative results on the real-world intent reasoning test set (88 samples). Progress score denotes the ratio of correctly predicted targets in each sample (e.g., a score of 50 indicates that half of the targets are correct).}
    \label{tab:intent_reasoning}
    \resizebox{0.485\textwidth}{!}{
    \begin{tabular}{lcc}
        \toprule
        Method & Acc (\%) & Progress Score (\%) \\
        \midrule
        Baseline-1 (prompted MLLM) & 38.6 & 61.4 \\
        Baseline-2 (geometric pipeline) & 59.1 & 78.4 \\
        \rowcolor{myblue}
        \textbf{GesVLA ($\text{VLM}_{\text{int}}$)} & \textbf{94.3} & \textbf{97.2} \\
        \bottomrule
    \end{tabular}
    }
    \vspace{-.1cm}
\end{table}

\textbf{Intent Reasoning Experiment.}
To validate sim-to-real transfer of gesture-conditioned intent reasoning, $\text{VLM}_{\text{int}}$ is trained on semi-synthetic data and directly evaluated in real-world scenarios. The model uses only the right-view camera, which provides a clear view of the pointing process.

The intent reasoning experiments are shown in Fig.~\ref{fig:setup}.
Several blocks and plates are placed on the table. A fixed instruction is used: “Pick this up and put it there.” We collect 300 RGB-D scenes to construct the semi-synthetic training dataset, and evaluate on 88 additional scenes with varying object numbers ($\ge 7$) and arrangements. In each test trial, an experimenter provides a gesture-language command, and the prediction is considered correct only if both the pointed block and the pointed plate are accurately predicted.

\textbf{Manipulation Experiment.}
We use a 7-DOF ARX5 robotic arm for real-world evaluation. 
All three cameras are enabled to provide a comprehensive observation view.
Three tasks are designed to validate the model’s ability to interpret multimodal instructions and to demonstrate its potential in daily-life scenarios, as shown in Fig.~\ref{fig:setup}. Each task is repeated 20 times, and the average success rate is reported. 

\begin{itemize}[leftmargin=11pt]
    \item \textbf{Pick-and-Place Block:} 
    The robot must grasp a specified block from a cluttered arrangement of blocks and place it onto the designated plate. A trial is counted as successful if the correct block is grasped and placed properly.
    
    \item \textbf{Select Jelly:} 
    Multiple jelly cups are placed on several plates on the table. The robot must grasp one or more specified jelly cups in the indicated pointing order and place them into a target plate. Success is defined as correctly grasping the target objects from the correct plates in the correct order and completing the placement.
    
    \item \textbf{Select Fruit and Vegetable:} 
    Various bell peppers and bananas are placed in two produce bins. The robot must grasp one or more specified objects in the indicated pointing order and place them into a basket. A trial is considered successful only if the exact target objects are grasped and placed in the correct order.
\end{itemize}

\subsection{Intent Reasoning Experiment Results}\label{reasoning_result}

According to the experimental setup (Sec.~\ref{exp_setup}), we test on 88 manually collected samples and conduct experiments with two baselines. Table~\ref{tab:intent_reasoning} summarizes the results.

\textbf{Baseline-1 (Prompted Native MLLM).} We input key gesture frames into Qwen3.5-plus~\cite{yang2025qwen3} and prompt the model to output the pointing coordinates for target determination. This method achieves an accuracy of 34/88 (38.6\% on the 88 samples). Error analysis shows that the model often selects the ``closest object to the finger'' rather than the ``true target along the pointing direction,'' even after we repeatedly optimized the prompts, indicating insufficient modeling of the geometric pointing relationship of the gesture. 

\textbf{Baseline-2 (Geometric Pipeline).} We implement a non-end-to-end pipeline using MediaPipe~\cite{zhang2020mediapipe} for finger keypoint extraction, GroundingDINO~\cite{liu2023grounding} for object detection, and DepthAnythingV2~\cite{depth_anything_v2} for depth estimation. The target is determined by reconstructing a sparse 3D point cloud from candidate bounding boxes and minimizing the distance to the index finger ray. This pipeline achieves 52/88 (59.1\%). Although it outperforms Baseline-1, its robustness is weak: errors from any submodule (keypoints, detection, depth) propagate and cause failure. Moreover, language and gesture are decoupled, limiting flexible user intention understanding.

\textbf{GesVLA.} Our method achieves the highest accuracy of 83/88 (94.3\%), outperforming Baseline-2 and Baseline-1 by +35.2\% and +55.7\%. The results show that joint modeling of gesture and language significantly improves the accuracy and robustness of intention understanding. Even when local hand observations are unstable, the model can still leverage hand pose context and language priors to complete intent reasoning. Moreover, joint modeling provides consistent intention representations for downstream policy generation, demonstrating greater practical deployment value.

\subsection{Manipulation Experiment Results}

We compare GesVLA against three baselines that represent alternative strategies for incorporating spatial guidance, and also report an intermediate variant of our own method to isolate the benefit of end-to-end feature interaction.

\textbf{Baseline-1 (Text-only VLA).} To isolate the contribution of gesture, we replace each gesture sequence in the training data with a descriptive paragraph specifying object appearance and positions, and fine-tune a similar VLA model~\cite{black2024pi_0}. This baseline achieves a total average success rate of only 31.7\%, confirming that language descriptions alone are insufficient for precise spatial grounding in cluttered scenes.

\textbf{Baseline-2 and 3 (Prompted Native MLLM + VLA and Geometric Pipeline + VLA).} We adopt the same prompted MLLM and geometric pipeline as in Sec.~\ref{reasoning_result} for intent reasoning, and use their predicted targets to generate visual prompts (highlighted points on target objects) for the VLA policy~\cite{black2024pi_0}. The real-robot training data and training setup are kept identical to ours for fair comparison. Both baselines perform poorly, achieving average success rates of 31.7\% and 41.7\%, respectively. The low accuracy directly inherits the weak gesture reasoning of their front-end modules.

\textbf{GesVLA (Decoupled VLMs).} Decoupling $\mathrm{VLM}_{\mathrm{int}}$ and $\mathrm{VLM}_{\mathrm{per}}$ by removing cross-attention while keeping the visual prompt interface raises the total average to 61.7\%. The remaining gap to the full model indicates that visual prompts still introduce a modality bottleneck.

\textbf{GesVLA.} Our full model outperforms all other variants by a large margin. The advantage is most evident in cluttered scenes, where the model reliably follows sequential pointing commands. These results demonstrate that end-to-end latent interaction between gesture perception and action generation is essential for robust spatial instruction following.

\subsection{Ablation Study}

\begin{table}[t]
    \small
    \centering
    \caption{Ablation on intent reasoning.}
    \vspace{-10pt}
    \label{tab:ablation_reasoning}
    \setlength{\tabcolsep}{4pt}
    \begin{tabular}{cc}
        \begin{subtable}{0.47\columnwidth}
            \centering
            \caption{Data engine operations.}
            \label{tab:ablation_data}
            \vspace{-4pt}
            \begin{tabular}{lcc}
                \toprule
                Variant & Acc.\ (\%) \\
                \midrule
                w/o coord.\ jitter & 42.0\\
                w/o hand augm. & 76.1 \\
                \rowcolor{myblue}
                \textbf{Full pipeline} & \textbf{94.3} \\
                \bottomrule
            \end{tabular}
        \end{subtable}
        &
        \begin{subtable}{0.47\columnwidth}
            \centering
            \caption{Training design.}
            \label{tab:ablation_train}
            \vspace{-4pt}
            \begin{tabular}{lcc}
                \toprule
                Variant & Acc.\ (\%) \\
                \midrule
                w/o gesture MLP & 84.1 \\
                w/o data augm. & 89.8 \\
                \rowcolor{myblue}
                \textbf{Full model} & \textbf{94.3} \\
                \bottomrule
            \end{tabular}
        \end{subtable}
    \end{tabular}
    \vspace{-.1cm}
\end{table}

\begin{table}[t]
    \small
    \centering
    \caption{Ablation on training strategy.}
    \label{tab:ablation_strategy}
    \vspace{-.15cm}
    \resizebox{0.485\textwidth}{!}{
    \begin{tabular}{lcccc}
        \toprule
        Variant & Block & Jelly & Fruit/Veg. & Avg (\%) \\
        \midrule
        Joint training (all modules)  & 10/20 & 8/20 & 9/20 & 45.0 \\
        Two stage ($\text{VLM}_{\text{int}}$ unfrozen)  & \textbf{20/20} & \textbf{15/20} & 13/20 & 80.0 \\
        \rowcolor{myblue}
        \textbf{Two stage ($\text{VLM}_{\text{int}}$ frozen)} & 19/20 & \textbf{15/20} & \textbf{16/20} & \textbf{83.3} \\
        \bottomrule
    \end{tabular}
    }
    \vspace{-.1cm}
\end{table}

\begin{table}[t]
    \small
    \centering
    \caption{Ablation on policy model architecture.}
    \label{tab:ablation_arch}
    \vspace{-.15cm}
    \resizebox{0.477\textwidth}{!}{
    \begin{tabular}{lcccc}
        \toprule
        Variant & Block & Jelly & Fruit/Veg. & Avg (\%) \\
        \midrule
        w/o visual prompt & 15/20 & 12/20 & 13/20 & 66.6 \\
        + text prompt & 18/20 & \textbf{17/20} & 15/20 & \textbf{83.3} \\
        \rowcolor{myblue}
        \textbf{Full model} & \textbf{19/20} & 15/20 & \textbf{16/20} & \textbf{83.3} \\
        \bottomrule
    \end{tabular}
    }
    \vspace{-.2cm}
\end{table}

\begin{figure}[t]
    \centering
    \includegraphics[width=0.92\linewidth]{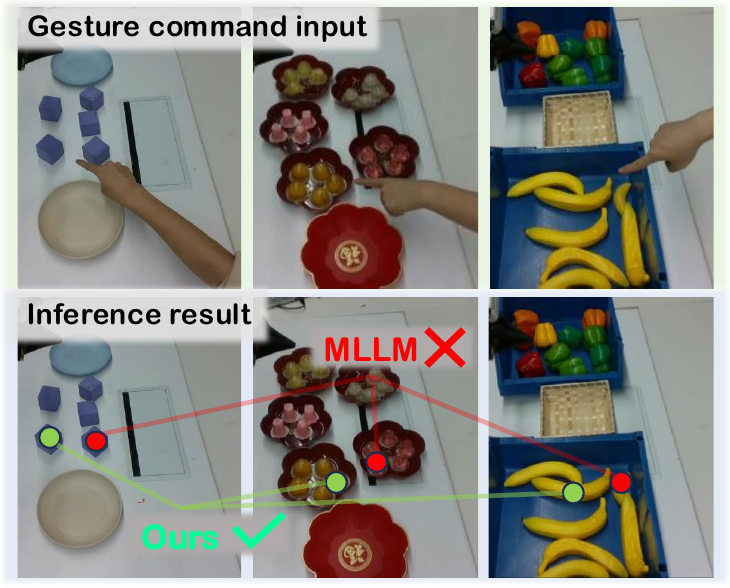}
    \vspace{-.15cm}
    \caption{
    Qualitative comparison of gesture-conditioned intent reasoning between a native MLLM and our $\text{VLM}_{\text{int}}$. 
    The native MLLM often selects objects closest to the fingertip in image space, while our method correctly infers the intended target along the pointing direction.
    }
    \label{fig:reasoning_vis}
    \vspace{-.3cm}
\end{figure}

\textbf{Intent Reasoning.}
As shown in Table~\ref{tab:ablation_data}, removing coordinate jitter during synthetic data generation causes a severe drop (42.0\%), because the model overfits to fixed coordinate bins and fails to generalize.
Removing hand appearance augmentation degrades performance to 76.1\%, confirming that diverse hand visual variations are crucial for sim-to-real transfer.
In Table~\ref{tab:ablation_train}, removing the gesture MLP projection layer reduces accuracy to 84.1\%. Although raw gesture images still provide partial spatial cues, the explicit pointing-direction information encoded by the gesture vector significantly improves target localization.
Training data augmentation (color jitter, cropping, translation) further improves robustness, and the full design achieves 94.3\%.

\textbf{Training Strategy.}
Table~\ref{tab:ablation_strategy} compares different training strategies for the dual-VLM policy.
Joint training optimizes the entire framework simultaneously using real-robot action supervision together with semi-synthetic gesture reasoning supervision, achieving only 45.0\% average success. This setting causes the large-scale synthetic reasoning data to dominate optimization and negatively affect downstream action learning.
The two-stage strategy first pre-trains $\mathrm{VLM}_{\mathrm{int}}$, and then trains $\mathrm{VLM}_{\mathrm{per}}$ and the action expert. Fine-tuning $\mathrm{VLM}_{\mathrm{int}}$ in the second stage with additional real-robot gesture annotations achieves 80.0\% average success, while freezing $\mathrm{VLM}_{\mathrm{int}}$ achieves the best performance of 83.3\% without requiring extra gesture supervision. This demonstrates that the semi-synthetic gesture data already provides sufficient supervision for effective sim-to-real transfer.

\textbf{Model Architecture.}
In Table~\ref{tab:ablation_arch}, removing the visual prompt significantly degrades overall success, demonstrating that visualized spatial hints are essential for target grounding.
Adding textual prompts from $\mathrm{VLM}_{\mathrm{int}}$ does not bring performance improvement, confirming that cross-attention already conveys intent information effectively.

\subsection{Qualitative Comparison}
\textbf{Real-world Rollout Visualization.}
We provide complete real-world rollout demonstrations for all three manipulation tasks in the supplementary video. We compare GesVLA with the text-only VLA baseline under the same task settings. The results show that introducing gesture as an additional instruction modality significantly simplifies human-robot interaction and improves the accuracy of intention understanding, especially in cluttered scenes requiring spatial disambiguation.

\textbf{Intention Understanding.}
We further provide qualitative comparisons between a native MLLM~\cite{yang2025qwen3} and our intent reasoning model $\text{VLM}_{\text{int}}$. As shown in Fig.~\ref{fig:reasoning_vis}, the evaluation examples are collected from three real-world tasks. Given the same gesture-language instruction, the native MLLM frequently fails to correctly infer the intended target object. Instead, it tends to select objects closest to the fingertip in image space, without explicitly reasoning about the pointing direction and spatial geometry.
In contrast, our method successfully identifies the intended targets along the pointing direction in cluttered scenes. The results demonstrate that gesture-aware intent reasoning substantially improves spatial grounding robustness and provides more reliable intention understanding for downstream robotic manipulation.

\section{CONCLUSION}
In this paper, we present GesVLA, a gesture-aware VLA framework for embodied manipulation. By introducing pointing gestures as an additional instruction modality, GesVLA effectively reduces spatial ambiguity in cluttered real-world environments. We propose a gesture-aware dual-VLM architecture with efficient asymmetric interaction, together with a scalable semi-synthetic gesture data engine and a two-stage training pipeline for intent reasoning and policy learning.
Experiments on real-world manipulation tasks demonstrate that GesVLA significantly improves both target grounding accuracy and downstream task success rates compared with text-only and MLLM-based baselines. Qualitative results further show that gesture-aware interaction enables more intuitive and reliable human-robot communication.

Nevertheless, the current system only considers widely used pointing gestures and does not yet support more challenging human-robot collaboration scenarios. In future work, we plan to explore richer human instruction modalities and extend GesVLA to diverse manipulation tasks.

\addtolength{\textheight}{-12cm}


\bibliographystyle{IEEEtran}
\bibliography{references}

\end{document}